\documentclass{amia}
\usepackage{lipsum} %Remove if not needed
\usepackage{multirow}
\usepackage{amssymb}
\usepackage{amsmath}
\usepackage{booktabs}
\usepackage{xcolor}
\usepackage{longtable,natbib}
\usepackage{wrapfig}
\usepackage{float}
\newcommand{\oms}[2]{{$#1$}}
\newcommand{\nms}[2]{{$#1$\tiny{$\pm$$#2$}}}

\usepackage{tikz}
\newcommand{\nerbox}[1]{\tikz[baseline=(X.base)]\node [draw=orange!50,fill=blue!15,rectangle,inner sep=1pt, rounded corners=3pt] (X) {#1};}

\begin{document}

\title{Does Synthetic Data Generation of LLMs Help Clinical Text Mining?}

\author{Ruixiang Tang$^1$*, Xiaotian Han$^2$*, Xiaoqian Jiang$^3$, Xia Hu$^1$}
\institutes{
    $^1$ Rice University;
    $^2$ Texas A\&M University;
    $^3$ University of Texas Health Science Center, School of Biomedical Informatics;
}

\maketitle

\section*{Abstract}
\textit{Recent advancements in large language models (LLMs) have led to the development of highly potent models like OpenAI's ChatGPT. These models have exhibited exceptional performance in a variety of tasks, such as question answering, essay composition, and code generation. However, their effectiveness in the healthcare sector remains uncertain. In this study, we seek to investigate the potential of LLMs to aid in clinical text mining by examining their ability to extract structured information from unstructured healthcare texts, with a focus on biological named entity recognition and relation extraction. However, our preliminary results indicate that employing LLMs directly for these tasks resulted in poor performance and raised privacy concerns associated with uploading patients' information to the LLM API. To overcome these limitations, we propose a new training paradigm that involves generating a vast quantity of high-quality synthetic data with labels utilizing LLMs and fine-tuning a local model for the downstream task. Our method has resulted in significant improvements in the performance of downstream tasks, improving the F1-score from 23.37\% to 63.99\% for the named entity recognition task and from 75.86\% to 83.59\% for the relation extraction task. Furthermore, generating data using LLMs can significantly reduce the time and effort required for data collection and labeling, as well as mitigate data privacy concerns. In summary, the proposed framework presents a promising solution to enhance the applicability of LLM models to clinical text mining.}

\section{Introduction}

Recent advances in large language models (LLM) have dramatically improved the performance of various natural language processing (NLP) tasks, creating new opportunities for automating tasks traditionally performed by humans. OpenAI's ChatGPT, for instance, has demonstrated its ability to perform as well as humans on MBA exams at the Wharton Business School \cite{chatgpt2023MBA}, showcasing its competitiveness with human knowledge and its potential to assist professionals \cite{choi2023chatgpt, baidoo2023education}. In the healthcare industry, LLMs hold tremendous potential for transforming the field by extracting valuable insights from unstructured data, such as electronic health records and digital medical data. By identifying crucial data points for population health management, clinical trials, and drug discovery, LLMs can help facilitate the development of new drugs and treatment plans. As the development and integration of LLMs in healthcare continue, professionals can expect significant improvements in patient outcomes and overall healthcare delivery. % Additionally, LLMs could help identify essential data points for population health management, clinical trials, and drug discovery, which can facilitate the development of new drugs and treatment plans. %The potential of LLMs to analyze large datasets with speed and accuracy can reduce the time and effort required for data collection, processing, and analysis. This can lead to significant improvements in patient care, disease diagnosis, and treatment outcomes. 

LLMs possess a distinct advantage in their emergent abilities in Zero-Shot Learning, enabling them to learn and adapt to new tasks through prompt instructions, even if they have never encountered them before \cite{wei2022emergent, liu2023pre}. For example, by incorporating the instruction prompt "Translate these sentences from [source Language] to [Target Language]:", ChatGPT can compete favorably with commercial translation products, such as Google Translate \cite{jiao2023chatgpt}. While LLM’s ability to conduct many NLP tasks makes it a valuable tool for users, it also raises significant privacy issues. One major concern is that sensitive information may be inadvertently revealed during the process. This is particularly true in healthcare, where the privacy and confidentiality of patient information are of utmost importance. Therefore, it is important to ensure that there is a mechanism to ensure robust privacy protections that prevent unauthorized access to sensitive information. In addition, reliability and usability are important issues that must be addressed when using LLM. Users must be able to rely on the accuracy and consistency of the model, which requires ongoing refining, testing, and evaluation to ensure that the system is functioning as intended and meeting the needs of users. In this paper, we will focus on improving the reliability of LLM for zero-shot tasks while mitigating the privacy risk.

% While prior studies primarily focused on traditional natural language understanding tasks, the zero-shot ability of LLMs in the healthcare domain has yet to be fully understood. But there are also unique challenges in privacy and security risks and reliability/usability issues that we need to address.

In order to assess the zero-shot performance of current LLM models for healthcare tasks, we conducted experiments on ChatGPT to investigate its ability to extract structured information from unstructured healthcare texts, specifically for biological named entity recognition (NER) and relation extraction (RE) tasks. Our preliminary findings suggest that using ChatGPT directly only yields poor performance compared to SOTA models trained on the dataset, as indicated in Table \ref{tab:ner_motivation} and \ref{tab:re_motivation}. This result highlights the fact that while ChatGPT has demonstrated impressive inference and reasoning abilities in various classic natural language understanding (NLU) tasks, it is not adequate to apply ChatGPT alone to healthcare tasks since it was not specifically trained for this domain \cite{brown2020language, qin2023chatgpt}. In addition to the performance limitations, integrating large language models into hospital systems raises privacy concerns as most LLMs are only available through their APIs, and healthcare providers cannot upload patient information directly to the LLMs' APIs \cite{aziz2017review}.

To bridge the gap, we propose a novel training paradigm to tackle the challenges of utilizing LLMs for healthcare tasks. Instead of directly applying LLMs in a zero-shot setting, we generate a large volume of synthetic data with labels using LLMs. To improve the quality and diversity of the synthetic data, we use a small amount of human-labeled examples as seeds and create appropriate prompts to guide LLMs in generating a variety of examples with varying sentence structures and linguistic patterns. A post-processing step is employed to eliminate low-quality or duplicated samples produced by LLMs. Finally, we employ the synthetic data to fine-tune a local pre-trained language model. Our experiments on four representative datasets show that the proposed pipeline significantly enhances the performance of the local model compared to LLMs' zero-shot performance. In addition, the local offline model effectively addresses data privacy concerns by reducing the need for uploading patient data to LLM APIs. In conclusion, our innovative training framework enables the training of a local model with superior performance compared to using LLMs alone. It also mitigates potential privacy concerns and reduces the dependence on costly and time-consuming data collection and labeling.

%Due to the sensitive and private nature of healthcare-related applications, there are potential problems associated with directly leveraging ChatGPT for these applications, including data privacy, such as the risk of exposing confidential patient information and the accuracy of the model's predictions in the zero-shot setting. In this section, we discuss the potential harms of directly using ChatGPT for healthcare tasks.

%\textbf{The dataset privacy concerns.} Another challenge with using ChatGPT for healthcare-related tasks is the privacy of patient data. Directly uploading the electronic health record (EHR) dataset of the patient to ChatGPT would raise serious privacy concerns. The EHR dataset contains sensitive information about the patient's health, such as their medical history, current conditions, and treatment plans. Healthcare providers and institutions are bound by strict privacy regulations, such as HIPAA, which require them to protect the confidentiality of patient data.

\section{Preliminaries}

 \textbf{Biomedical Named Entity Recognition.} Biomedical NER involves identifying and categorizing medical entities, such as diseases, symptoms, drugs, etc., in a medical text. NER uses an IOB (Inside, Outside, Begin) tagging scheme, where each word is assigned a tag indicating whether it is the beginning of a named entity (B), inside a named entity (I), or outside a named entity (O). For example, the sentence "The symptoms suggest a possible case of rheumatoid arthritis." would be tagged as "O O O O O O O B-Disease I-Disease O". Formally, a sentence $s$ in a medical text is denoted as a sequence of words $\textbf{s} = (w_1, w_2, \cdots, w_n)$, and the corresponding tags for each word in the sentence are denoted as $\textbf{y} = (y_1, y_2, \cdots, y_n)$, where tag $y_i$ is an element of the tag set \{B, I, O\}. %Biomedical NER has versatile usage for the healthcare domain, including analyzing Electronic Health Records (EHRs)~\cite{gorinski2019named,dai2019named,kormilitzin2021med7}, extracting clinical trials~\cite{kang2017eliie,nye2021understanding}, and drug development~\cite{shinozaki2020electronic,rocktaschel2012chemspot,eltyeb2014chemical}.

 \textbf{Biomedical Relation Extraction.} Biomedical RE involves identifying and extracting the relationships between medical entities in a text, such as diseases and drugs, symptoms and treatments, etc. Formally, let $x$ be a sentence containing two medical entities $e_1$ and $e_2$, and let $r$ be the relation between them. The MRE task can be formulated as a classification problem, where the goal is to learn a function $f(x, e_1, e_2) \rightarrow r, r \in \mathcal{R}$ that utilizes the context in the sentence $x$ to predict the relation between $(e_1, e_2)$. The performance of the medical NER and RE models is typically evaluated using standard classification task metrics such as precision, recall, and F1-score. 

\textbf{Zero-shot Learning.} Zero-shot learning is an emerging research paradigm that allows LLMs to perform tasks they have not been explicitly trained. This is accomplished by utilizing the LLMs' capacity to produce coherent text based on a given prompt. The prompt serves as a guide, providing a corpus that describes the task at hand along with a set of potential outputs \cite{liu2023pre}. The LLM then generates the most plausible output based on its acquired knowledge. Recent studies show that LLMs achieve promising zero-shot ability in various traditional NLU tasks \cite{zhong2023can}.

\textbf{Experimental Dataset.} For the NER task, we consider two widely used datasets, including National Center for Biotechnology Information disease corpus (NCBI) and the BioCreative V CDR corpus (BC5CDR)~\cite{li2016biocreative}. The NCBI dataset contains $6,881$ human-labeled annotations and is used to recognize disease names \cite{dougan2014ncbi}. The BC5CDR corpus collects $1,500$ PubMed articles with $4,409$ annotated chemicals, $5,818$ diseases, and $3,116$ chemical-disease interactions, and it is used in our study to evaluate the chemical and disease recognition task. For the RE task, we adopt two binary relation extraction datasets. The Gene Associations Database (GAD) dataset \cite{rouillard2016harmonizome}, which is a corpus of gene-disease associations curated from genetic association studies and contains 5,330 annotations. We also consider the EU-ADR corpus, a biomedical relation extraction dataset that contains 100 abstracts with relations between drugs, disorders, and targets \cite{van2012eu}. However, GAD and EU-ADR have noisy labels and are considered weakly supervised datasets. To ensure an accurate evaluation of the model's performance, we employed three annotators to manually label 200 data samples from the original test datasets for both GAD and EU-ADR. The ground truth label was determined through majority voting.

\section{Benchmarking LLM on Biomedical NER and RE Tasks}
We conducted benchmark experiments on
the ChatGPT model. To create prompts that would effectively trigger ChatGPT's named entity recognition and relation extraction abilities, we take inspiration from the ChatGPT itself by prompting it for advice. Specifically, we asked ChatGPT to provide us with five concise templates that could be used to address biological NER and RE tasks. After testing the generated prompts on a validation set, we selected the most effective ones and added instructions for adapting them to suit downstream datasets, prompts are shown in Table~\ref{tab: prompts}. 

\begin{table}[H]
    \centering
    \caption{Prompts for Named Entity Recognition Task and Relation Extraction Task}
    \begin{tabular}{l|p{0.4\textwidth}|p{0.4\textwidth}} 
    \toprule
        Task & Named Entity Recognition (NCBI Disease) & Relation Extraction (GAD Dataset)  \\ \midrule
        Prompt &  Please do NER task for "@TEXT" (output IOB format, please output the results only without your explanation, use tab key to separate the word and label, the entity is disease name, please use the space key to separate the sentences)
        
        & Given a sentence that introduces a gene (denoted as "@GENE\$") and a disease (denoted as "@DISEASE\$"), predict whether the gene and disease have a relation or not. The relation between the gene and disease can be any functional, causal, or associative connection. If there is a relation, then the label should be "Yes", otherwise "No". \\ \bottomrule
    \end{tabular}
    \label{tab: prompts}
\end{table}

In Tables~\ref{tab:ner_motivation} and~\ref{tab:re_motivation}, we report the performance of ChatGPT using our prompts and state-of-the-art models trained on the dataset. Results demonstrate that although ChatGPT shows some capability as a generalist model that can perform multiple traditional natural language understanding tasks \cite{guo2023close}, it performs inferior to SOTA models that have been fine-tuned on specific healthcare tasks. Specifically, ChatGPT performs slightly worse than SOTA in the biological relation extraction task, but there is a substantial performance gap between ChatGPT and SOTA on the biological named entity recognition task. For instance, the average disease recognition F1-score of the SOTA model is 88.60\%, whereas ChatGPT achieves only 35.93\%. Similarly, the average relation extraction F1-score of the SOTA model is 84.35\%, while ChatGPT achieves 78.35\%. This outcome is unsurprising, as ChatGPT is trained to tackle general natural language problems and has not been trained specifically for these tasks.

\begin{minipage}{0.495\textwidth}
\begin{table}[H]
\fontsize{8}{9}\selectfont  
\setlength{\tabcolsep}{4pt}
\centering 
\caption{NER Test Results of ChatGPT versus SOTA} \label{tab:ner_motivation}
\begin{tabular}{lccccccccc}
\toprule
            &Metrics  & SOTA       & ChatGPT &  Decrease \\
\midrule
% $\downarrow$
\multirow{3}{*}{NCBI Disease}       &P          & $82.87$ & $32.84$ & {\color{red}$\downarrow60.27\%$}  \\
                                    &R          & $89.54$ & $44.86$ & {\color{red}$\downarrow49.91\%$}  \\
                                    &F          & $86.08$ & $37.92$ & {\color{red}$\downarrow55.94\%$}  \\ \midrule
\multirow{3}{*}{BC5CDR Chemical}    &P          & $91.07$ & $ 5.76$ & {\color{red}$\downarrow94.36\%$}  \\
                                    &R          & $92.24$ & $11.69$ & {\color{red}$\downarrow87.31\%$}  \\
                                    &F          & $91.65$ & $ 7.72$ & {\color{red}$\downarrow91.58\%$}  \\ 
\bottomrule
\end{tabular}
\end{table}
\end{minipage}
\hfill
\begin{minipage}{0.495\textwidth}
\begin{table}[H]
\fontsize{8}{9}\selectfont  
\setlength{\tabcolsep}{4pt}
\centering 
\caption{RE Test Results of ChatGPT versus SOTA} \label{tab:re_motivation}
\begin{tabular}{lccccccccc}
\toprule
            &Metrics  & SOTA       & ChatGPT &  Decrease ($\downarrow$)\\
\midrule

\multirow{3}{*}{GAD}                &P          & $84.28$ & $76.32$ & {\color{red}$\downarrow7.96\%$}  \\
                                    &R          & $94.21$ & $79.82$ & {\color{red}$\downarrow14.39\%$}  \\
                                    &F          & $88.96$ & $78.03$ & {\color{red}$\downarrow10.93\%$}  \\ \midrule
\multirow{3}{*}{EU-ADR}             &P          & $75.81$ & $72.01$ & {\color{red}$\downarrow3.80\%$}  \\
                                    &R          & $81.20$ & $75.43$ & {\color{red}$\downarrow5.77\%$}  \\
                                    &F          & $78.41$ & $73.68$ & {\color{red}$\downarrow4.73\%$}  \\ 
\bottomrule
\end{tabular}
\end{table}
\end{minipage}

\section{Exploring Synthetic Data Generation of ChatGPT for Clinical Text Mining}\label{sec:meth}

\textbf{Motivation.} In Section 3, we demonstrate that ChatGPT achieves only average performance for the biomedical relation extraction task and poor performance for the biomedical named entity recognition task. Additionally, uploading patient data directly would pose significant privacy concerns. Regulations such as GDPR \cite{voigt2017eu} and CCPA \cite{goldman2020introduction} prohibit the upload of electronic health record information to ChatGPT, as this could potentially compromise patients' protected health information (PHI). To leverage ChatGPT's capabilities in assisting with healthcare-related tasks, we propose utilizing ChatGPT to generate a large volume of training data along with corresponding labels, which can be used to train a local model. This approach requires only a few examples to generate the entire training dataset, thereby resolving the low-resource problem commonly encountered in the healthcare domain. Additionally, the local model can address privacy concerns since the synthetic data contains no patient-sensitive information. Hospitals can use the local model instead of the ChatGPT to perform the downstream tasks while maintaining the privacy of patient data.

% \textbf{The zero-shot accuracy of ChatGPT for NER and RE are low }.
% \textbf{The dataset privacy concerns}. Directly uploading the data of patients would suffer privacy issues.  We can not upload the EHR dataset of the patient to ChatGPT. Therefore, to use ChatGPT to assist the healthcare-related tasks, we find alternative ways to address these two challenges. We aim to improve the zero-shot performance of ChatGPT for NER and RE tasks, as well as to ensure the privacy of patient data while using ChatGPT.

\begin{table}[H]
    \centering
    \fontsize{9}{15}\selectfont  
    \caption{Generated Texts and Labels for Named Entity Recognition Task. Target entities are highlighted.}
    % \begin{tabular}{p{0.4\textwidth}p{0.4\textwidth}} 
    \begin{tabular}{p{0.97\textwidth}} 
    \toprule
         %  \multicolumn{1}{c}{NCBI Disease} \\ \hline
         % \multicolumn{1}{c|}{The symptoms of AGU may include seizures, \colorbox{blue!40}{hearing loss}, and joint problems.}         \\ \hline
         % \multicolumn{1}{c|}{Text}  \\ \hline
         % \multicolumn{1}{c|}{Text}  \\ \hline
         % \multicolumn{2}{c}{BC5CDR Disease} \\ \hline
         % \multicolumn{1}{c|}{Text} \\ \hline
         % \multicolumn{1}{c|}{Text} \\ \hline
         % \multicolumn{1}{c|}{Text} \\ \bottomrule

        \multicolumn{1}{c}{NCBI \nerbox{Disease}}\\ \midrule
        The symptoms of AGU may include seizures, \nerbox{hearing loss}, and joint problems.       \\ \hline
        The diagnosis of \nerbox{Autoimmune lymphoproliferative syndrome} is based on clinical, laboratory, and genetic findings .\\ \hline
        \nerbox{AS} has been found to play a role in the development of inflammation and \nerbox{autoimmune disorders}. \\ \hline
        Mutations in the BRCA1 and BRCA2 genes are responsible for the majority of cases of hereditary breast and \nerbox{ovarian cancer}. \\ \hline
        Hereditary colorectal cancer syndrome \nerbox{familial adenomatous polyposis} is a rare genetic condition that causes the development of multiple \nerbox{polyps} in the colon and rectum. \\\midrule
         \multicolumn{1}{c}{BC5CDR \nerbox{Chemical}}\\ \midrule
         A meta-analysis of clinical trials found that "\nerbox{modafinil}" significantly improves cognitive performance in healthy individuals. \\ \hline
        \nerbox{Mesna} has been shown to effectively reduce the incidence of hemorrhagic cystitis in patients receiving \nerbox{cyclophosphamide}.\\ \hline
        The use of \nerbox{atropine} in conjunction with oximes remains the mainstay of therapy for organophosphorus ( OP ) poisons. \\ \hline
        The safety and tolerability of \nerbox{metoprolol} were assessed in a large, randomized, double-blind trial .\\ \hline
    \end{tabular}
    \label{tab: NER Generated Dataset}
\end{table}

\begin{table}[H]
    \centering
    \fontsize{9}{13}\selectfont  
    \caption{Generated Texts and Labels for Relation Extraction Task. The label indicates whether there is a relation between the target gene (denoted as "@GENE\$") and the disease (denoted as "@DISEASE\$"). }
    \begin{tabular}{p{0.90\textwidth}|p{0.05\textwidth}} 
    \toprule
          \multicolumn{1}{c|}{Texts} & \multicolumn{1}{c}{Labels} \\ \midrule
          \multicolumn{2}{c}{GAD Dataset} \\ \hline
         The study demonstrates that the @GENE\$ gene is directly linked to @DISEASE\$ development, with patients carrying the C/T variant showing a significantly higher risk. & \multicolumn{1}{c}{Yes} \\ \hline
         The results of our study suggest that @GENE\$ is involved in the etiology of @DISEASE\$, and highlight the potential for @GENE\$ as a target for the development of new treatments for this debilitating disease. & \multicolumn{1}{c}{Yes} \\ \hline
         Despite extensive research, no significant association was found between @GENE\$ and @DISEASE\$, indicating that other genetic or environmental factors may play a role in the development of this disease. & \multicolumn{1}{c}{No} \\ \hline
        This research demonstrates that mtDNA @GENE\$ is a fast and reliable method for detecting mutations, but does not support a role for the T3 394C @DISEASE\$ in ND1 gene in the pathogenesis of mitochondrial diabetes. & \multicolumn{1}{c}{No} \\ \hline
         \multicolumn{2}{c}{EU-ADR} \\ \hline
         The K121Q polymorphism in the ectonucleotide pyrophosphatase (@GENE\$) gene and the rs7566605 genotype located near insulin-induced gene 2 have been shown to be associated with @DISEASE\$ and obesity. & \multicolumn{1}{c}{Yes} \\ \hline
        The @GENE\$ gene has been found to play a role in the immune response, but its relationship with @DISEASE\$ has yet to be fully understood. More research is needed. & \multicolumn{1}{c}{Yes} \\ \hline
         SNPs in the intracellular pattern recognition receptor nucleotide-binding oligomerization domain-containing protein (@GENE\$, nucleotide oligomerization domain 2) are not associated with @DISEASE\$.  & \multicolumn{1}{c}{No} \\ \hline
         Our results showed that @GENE\$ polymorphism was not associated with susceptibility to @DISEASE\$, and therefore may not play a significant role in the development of the disease. & \multicolumn{1}{c}{No} \\ \bottomrule
    \end{tabular}
    \label{tab: RE Generated Dataset}
\end{table}

\begin{figure}[t]
  \centering
    \includegraphics[width=1.0\textwidth]{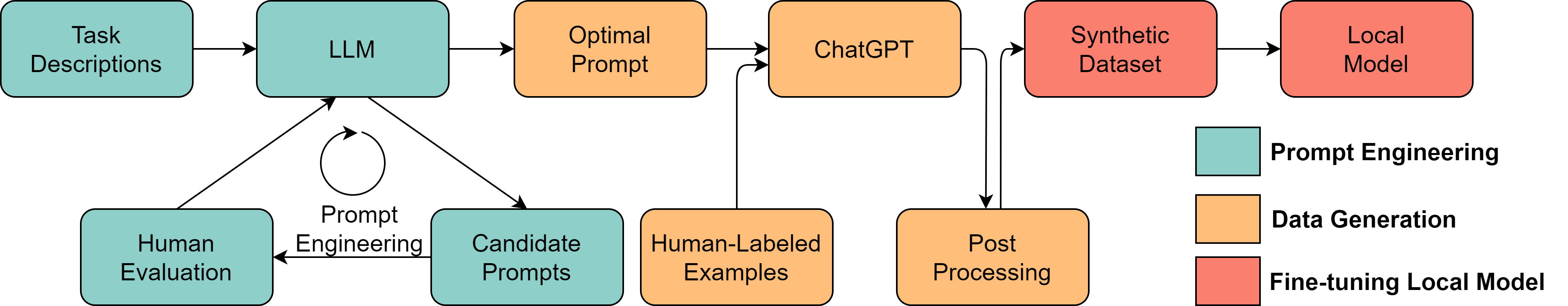}
 \caption{An overview of the workflow for synthetic data generation using ChatGPT.}
 \label{fig:workflow}
\end{figure}

\textbf{Prompt Engineering and Data Generation.} To guide ChatGPT in generating the synthetic dataset for our tasks, we designed a suitable instruction prompt inspired by ChatGPT itself. Our approach involved prompting ChatGPT with the following request: “Provide five concise prompts or templates that can be used to generate data samples of [Task Descriptions].” ChatGPT then provided us with five candidate prompts for data generation. We generated 10 data samples using each prompt and manually compared their quality to select the best prompt. As shown in Figure \ref{fig:workflow}, we repeated this process, asking ChatGPT to augment five prompts based on the previous best prompt until we arrived at the optimal prompt for data generation after three rounds of testing.

We present the optimal prompts for data generation in Table \ref{tab: generation prompts}. To ensure the synthetic data's high quality, we adopt the following measures: (1) To ensure the generated data is in the same distribution as the target dataset, we instruct ChatGPT to generate sentences that mimic the style of PubMed Journal articles, which is the data source of the NCBI and GAD dataset. (2) To prevent ChatGPT from generating duplicated examples, we added different seeds in the prompt for each round of data generation. For named entity recognition, we used different entity seeds, e.g., "familial adenomatous polyposis," and asked ChatGPT to generate sentences containing the seed entity.  For the relation extraction task, we use examples from the original dataset combined in a format of $\vert$ sentence $\vert$ label $\vert$ as seeds. As we previously mentioned, since the GAD and EU-ADR datasets are highly noisy, we manually selected 50 positive and 50 negative samples from the original training dataset. For each round, we randomly sampled three positive examples and three negative examples as the seed samples. The generated examples, as shown in Table \ref{tab: NER Generated Dataset} and \ref{tab: RE Generated Dataset}, are fluent and similar to sentences sampled from scientific articles.

\begin{table}[h]
    \centering
    \caption{Prompts for Named Entity Recognition Task and Relation Extraction Task}
    \begin{tabular}{l|p{0.4\textwidth}|p{0.45\textwidth}} 
    \toprule
        Task & Named Entity Recognition (NCBI Disease) & Relation Extraction (GAD Dataset) \\ \midrule
        Prompt &  Please act as a sentence generator for the biological domain and provide 30 sentences containing the words [Seed Entities]. These sentences should not include any additional information or explanation. Generated sentences should mimic the style of PubMed journal articles, using a variety of sentence structures:
        
        & Generate 3 positive and 3 negative examples for the gene-disease relation extraction task. The target gene is denoted as "@GENE\$" and the target disease is denoted as "@DISEASE\$".  The label is whether there is a relation between the target gene and disease. The relationship can be any functional, causal, or associative connection. If there is a relation, then the label should be "Yes". If there is no relation, the label should be "No". Sentences mimic the style of PubMed journal articles with various sentence structures. [Seed Examples]: \\ \bottomrule
    \end{tabular}
    \label{tab: generation prompts}
\end{table}

\section{Named Entity Recognition}
In this section, we evaluate the effectiveness of our proposed synthetic data generation approach for the named entity recognition (NER) task, following the methodology outlined in Section~\ref{sec:meth}. We first extract the $M$ seed entities from the training set and use them to generate synthetic sentences with annotations for the target entity type. Specifically, for each seed entity, we generate $N$ sentences with the corresponding entity annotations, in our experiments, we set $N=30$. We then use the synthetic dataset to fine-tune three pre-trained language models. To evaluate the performance of the baseline models and our proposed methods, we use a subset of the test set for all datasets due to the computational limitations of ChatGPT. We report the precision, recall, and F1 scores for each model. The models evaluated in our experiments include three settings: (1) zero-shot, where the models were not fine-tuned on any dataset, and the model is ChatGPT. (2) models fine-tuned on synthetic data generated by our approach, and (3) models fine-tuned on the original training set. The pre-trained language models used in our experiments are BERT~\cite{devlin2018bert}, RoBERTa~\cite{liu2019roberta}, and BioBERT~\cite{lee2020biobert}.

\paragraph{Main Results.} We present the comparison of the baseline methods and our methods in Table~\ref{tab:ner}. From the experimental results, we observed that fine-tuning the models on the synthetic data generated using our approach leads to significant improvements compared to the zero-shot scenario in all the evaluated metrics. The average performance of BERT fine-tuned on synthetic data improved more than $35\%$ on Precision, $34\%$ on Recall, and $36\%$ on F1 than ChatGPT. Moreover, in some cases, our proposed method even achieves comparable performance to the models fine-tuned on the original training set. For example, for the BC5CDR Chemical dataset, the Recall of the BERT model fine-tuned on synthetic data obtained $81.96\%$, improved from $11.69\%$ in the zero-shot scenario, which is comparable to  the recall $88.66\%$ when fine-tuned on the original training set. The results demonstrate the effectiveness of our synthetic data generation approach in improving the performance of these models.

\vspace{10pt}
\begin{table}[h]
\fontsize{8}{10}\selectfont  
\setlength{\tabcolsep}{7pt}
\centering 
\caption{Test results in biomedical named entity recognition. Precision (P), Recall (R), and F1 (F) scores on each dataset are reported. All the numbers are in percentage and computed based on $3$ trials.} \label{tab:ner}
\vspace{-10pt}
\begin{tabular}{lccccccccc}
\toprule
&               &\multicolumn{1}{c}{Zero-shot}    &\multicolumn{3}{c}{Fine-Tuned on Synthetic Data} & \multicolumn{3}{c}{Fine-Tuned on Original Data}\\
\cmidrule(lr){3-3}\cmidrule(lr){4-6}\cmidrule(lr){7-9}
&Metrics    & ChatGPT           & BERT                 & RoBERTa             & BioBERT              & BERT      & RoBERTa              & BioBERT \\
\midrule

\multirow{3}{*}{NCBI Disease}       &P         & \oms{32.84}{00.00}  & \nms{39.41}{ 0.11}  & \nms{42.83}{ 0.48}  & \nms{43.14}{ 0.18}   & \nms{80.39}{ 1.70}   & \nms{84.62}{ 1.16}  & \nms{82.87}{ 1.50}     \\
&R         & \oms{44.86}{00.00}  & \nms{59.15}{ 0.53}  & \nms{62.78}{ 2.37}  & \nms{63.92}{ 0.41}   & \nms{86.18}{ 0.23}   & \nms{87.32}{ 0.53}  & \nms{89.54}{ 1.04}     \\
&F         & \oms{37.92}{00.00}  & \nms{47.30}{ 0.09}  & \nms{50.91}{ 1.10}  & \nms{51.51}{ 0.22}   & \nms{83.18}{ 0.86}   & \nms{85.94}{ 0.35}  & \nms{86.08}{ 1.29}     \\ \midrule
\multirow{3}{*}{BC5CDR Disease}     &P         & \oms{17.03}{00.00}  & \nms{62.51}{ 0.40}  & \nms{64.47}{ 0.59}  & \nms{63.08}{ 0.68}   & \nms{71.24}{ 1.39}   & \nms{80.70}{ 1.88}  & \nms{76.96}{ 2.31}     \\
&R         & \oms{43.56}{00.00}  & \nms{61.85}{ 0.08}  & \nms{62.95}{ 0.18}  & \nms{64.63}{ 0.59}   & \nms{79.24}{ 0.62}   & \nms{83.21}{ 0.37}  & \nms{84.78}{ 1.22}     \\
&F         & \oms{24.48}{00.00}  & \nms{62.18}{ 0.16}  & \nms{63.70}{ 0.19}  & \nms{63.84}{ 0.31}   & \nms{75.02}{ 1.04}   & \nms{81.93}{ 1.13}  & \nms{80.66}{ 0.71}     \\ \midrule  
\multirow{3}{*}{BC5CDR Chemical}    &P         & \oms{ 5.76}{00.00}  & \nms{62.45}{ 2.42}  & \nms{67.56}{ 0.84}  & \nms{68.88}{ 0.83}   & \nms{87.10}{ 1.62}   & \nms{91.12}{ 0.93}  & \nms{91.07}{ 0.19}     \\
&R         & \oms{11.69}{00.00}  & \nms{81.96}{ 1.89}  & \nms{83.36}{ 1.06}  & \nms{86.36}{ 0.76}   & \nms{88.66}{ 2.02}   & \nms{90.62}{ 0.50}  & \nms{92.24}{ 0.46}     \\
&F         & \oms{ 7.72}{00.00}  & \nms{70.84}{ 0.95}  & \nms{74.63}{ 0.81}  & \nms{76.64}{ 0.78}   & \nms{87.87}{ 0.72}   & \nms{90.87}{ 0.58}  & \nms{91.65}{ 0.21}     \\ \midrule
\multirow{3}{*}{Average}            &P         & \oms{18.54}{00.00}  & \oms{54.79}{00.00}  & \oms{58.28}{00.00}  & \oms{58.36}{00.00}   & \oms{79.57}{00.00}   & \oms{85.48}{00.00}  & \oms{83.63}{00.00}     \\
&R         & \oms{33.37}{00.00}  & \oms{67.65}{00.00}  & \oms{69.69}{00.00}  & \oms{71.63}{00.00}   & \oms{84.69}{00.00}   & \oms{87.05}{00.00}  & \oms{88.85}{00.00}     \\
&F         & \oms{23.37}{00.00}  & \oms{60.10}{00.00}  & \oms{63.08}{00.00}  & \oms{63.99}{00.00}   & \oms{82.02}{00.00}   & \oms{86.24}{00.00}  & \oms{86.13}{00.00}     \\
\bottomrule
\end{tabular}
\end{table}

\paragraph{The Effect of the Number of the Generated Sentences.} 
\begin{wrapfigure}[12]{R}{0.5\textwidth}
% \begin{figure}
    \centering
    \includegraphics[width=0.5\textwidth]{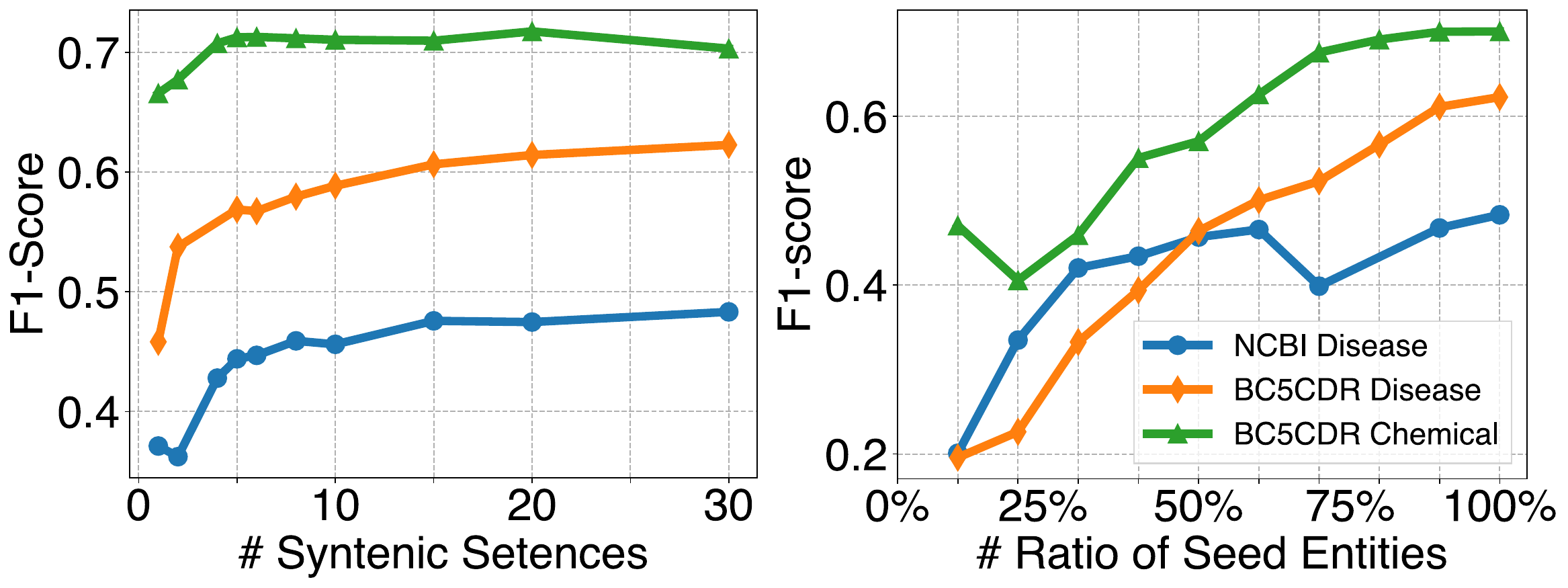}
    \caption{The performance with different numbers of synthetic sentences and the ratio of the seed entities. }
    \label{fig:effect_num}
% \end{figure}
\end{wrapfigure}
To investigate the impact of the number of synthetic sentences generated on the effectiveness of our proposed method, we conducted experiments with varying numbers of synthetic sentences and ratios of seed entities. As mentioned previously, we generated $N$ sentences with annotations for $M$ seed entities. In the first experiment, we used $10\%-90\%$ seed entities for synthetic data generation, while in the second experiment, we generated $[1,2,3,4,5,10,15,20,25,30]$ sentences for each entity. The results are presented in Figure~\ref{fig:effect_num}. Our findings showed that increasing the number of synthetic sentences can improve model performance up to a certain point, beyond which the improvement becomes marginal. Similarly, adjusting the ratio of synthetic to real entities in the training dataset can enhance model performance, especially for under-represented entities. %However, it's important to note that the effectiveness of using synthetic data is heavily dependent on the quality and diversity of the generated corpus and may not always lead to performance improvements.

% \paragraph{Case Study.} We conducted experiments with varying numbers of synthetic sentences and varying ratios of the entities in the training dataset.

\subsection{Relation Extraction}

In this section, we aim to assess the efficacy of our synthetic data generation approach for relation extraction. We follow the methodology outlined in Section~\ref{sec:meth} and randomly sample three positive and three negative examples from the 100 training manually labeled dataset to use as seed examples. For each round of generation, we generate three positive sentences and three negative sentences. In this way, we collect 6437 and 6424 synthetic examples for the GAD and EU-ADR, respectively. Given that the GAD and EU-ADR datasets are noisy, we manually label 200 test samples from the original test dataset to evaluate our models. We measure models' performance based on their precision, recall, and F1 scores. Our experiments cover three model settings: (1) zero-shot, where we directly leverage ChatGPT with the prompt shown in Table \ref{tab: prompts} for inference, (2) models fine-tuned on synthetic data generated by our approach, and (3) models fine-tuned on the original training set. Similarly to the NER task, we adopt BERT, RoBERTa, and BioBERT as the backbone models for our experiments.

% \vspace{10pt}
\begin{table}[h]
\fontsize{8}{10}\selectfont  
\setlength{\tabcolsep}{7pt}
\centering 
\caption{Test results in biomedical relation extraction. Precision (P), Recall (R), and F1 (F) scores on each dataset are reported. All the numbers are in percentage and computed based on 3 trials.}
\vspace{-10pt}
\begin{tabular}{lccccccccc}
\toprule
&               &\multicolumn{1}{c}{Zero-shot}    &\multicolumn{3}{c}{Fine-Tuned on Synthetic Data} & \multicolumn{3}{c}{Fine-Tuned on Original Data}\\
\cmidrule(lr){3-3}\cmidrule(lr){4-6}\cmidrule(lr){7-9}
&Metrics    & ChatGPT           & BERT                 & RoBERTa             & BioBERT              & BERT      & RoBERTa              & BioBERT \\
\midrule

\multirow{3}{*}{GAD}       &P         & \oms{76.32}{00.00}  & \nms{82.39}{ 0.93}   & \nms{83.59}{ 1.01}  & \nms{84.28}{ 1.03} & \nms{82.81}{ 0.12}  & \nms{83.75}{ 0.34}  & \nms{84.14}{ 0.12}     \\
&R         & \oms{79.82}{00.0}   & \nms{90.21}{ 0.15}   & \nms{92.57}{ 0.47}  & \nms{94.21}{ 1.35} & \nms{89.64}{ 0.21}  & \nms{91.32}{ 1.03}  & \nms{92.53}{ 0.89}    \\
&F         & \oms{78.03}{00.00}    & \nms{86.12}{ 0.72}   & \nms{87.85}{ 0.68}  & \nms{88.96}{ 1.01} & \nms{86.89}{ 0.15}  & \nms{87.37}{ 0.82}  & \nms{88.13}{ 0.31}    \\ \midrule
\multirow{3}{*}{EU-ADR}     &P         & \oms{72.01}{00.00}   & \nms{72.05}{ 1.02}   & \nms{73.44}{ 1.07}  & \nms{75.81}{ 1.43} & \nms{78.35}{ 0.21}  & \nms{79.88}{ 0.43}  & \nms{80.01}{ 0.57}    \\
&R         & \oms{75.43}{00.00}   & \nms{78.13}{ 0.50}   & \nms{79.22}{ 0.22}  & \nms{81.20}{ 1.00}  & \nms{84.61}{ 0.13}  & \nms{85.21}{ 0.20}  & \nms{87.81}{ 0.53}   \\
&F         & \oms{73.68}{00.00}   & \nms{74.96}{ 0.81}   & \nms{76.22}{ 0.55}  & \nms{78.41}{ 0.77}  & \nms{81.35}{ 0.15}  & \nms{82.46}{ 0.37}  & \nms{83.73}{ 0.24}   \\ \midrule  
\multirow{3}{*}{Average}            &P         & \oms{74.16}{00.00}     & \oms{77.22}{00.00}   & \oms{78.52}{00.00}  & \oms{80.05}{00.00}  & \oms{80.58}{00.00}  & \oms{81.82}{00.00}  & \oms{82.07}{00.00}   \\
&R         & \oms{77.62}{00.00}   & \oms{84.15}{00.00}   & \oms{85.90}{00.00}  & \oms{87.70}{00.00}   & \oms{87.13}{00.00}  & \oms{88.26}{00.00}  & \oms{90.17}{00.00}  \\
&F         & \oms{75.86}{00.00}   & \oms{80.53}{00.00}   & \oms{82.04}{00.00}  & \oms{83.69}{00.00}   & \oms{84.12}{00.00}  & \oms{84.91}{00.00}  & \oms{85.93}{00.00}   \\
\bottomrule
\end{tabular}
\label{fig: RE fine-tune}
\end{table}

\paragraph{Main Results.} We present the comparison of the baseline methods and our methods in Table~\ref{fig: RE fine-tune}. From the experimental results, we observed that fine-tuning the models on the synthetic data generated using our approach leads to notable improvements compared to the zero-shot scenario in all the evaluated metrics. The average performance on Precision, Recall, and F1 improve more than 6\%, 10\%, and 8\% percentage than ChatGPT. The model trained on the synthetic data achieves comparable performance to the models fine-tuned on the original training set. Notably, for the GAD dataset, the model trained on the synthetic dataset achieved slightly better results than the original dataset. The results demonstrate the effectiveness of our synthetic data generation approach for the relation extraction task.

\paragraph{The Effect of the Number of the Generated Sentences.} 
\begin{wrapfigure}[12]{R}{0.5\textwidth}
% \begin{figure}
    \centering
    \includegraphics[width=0.5\textwidth]{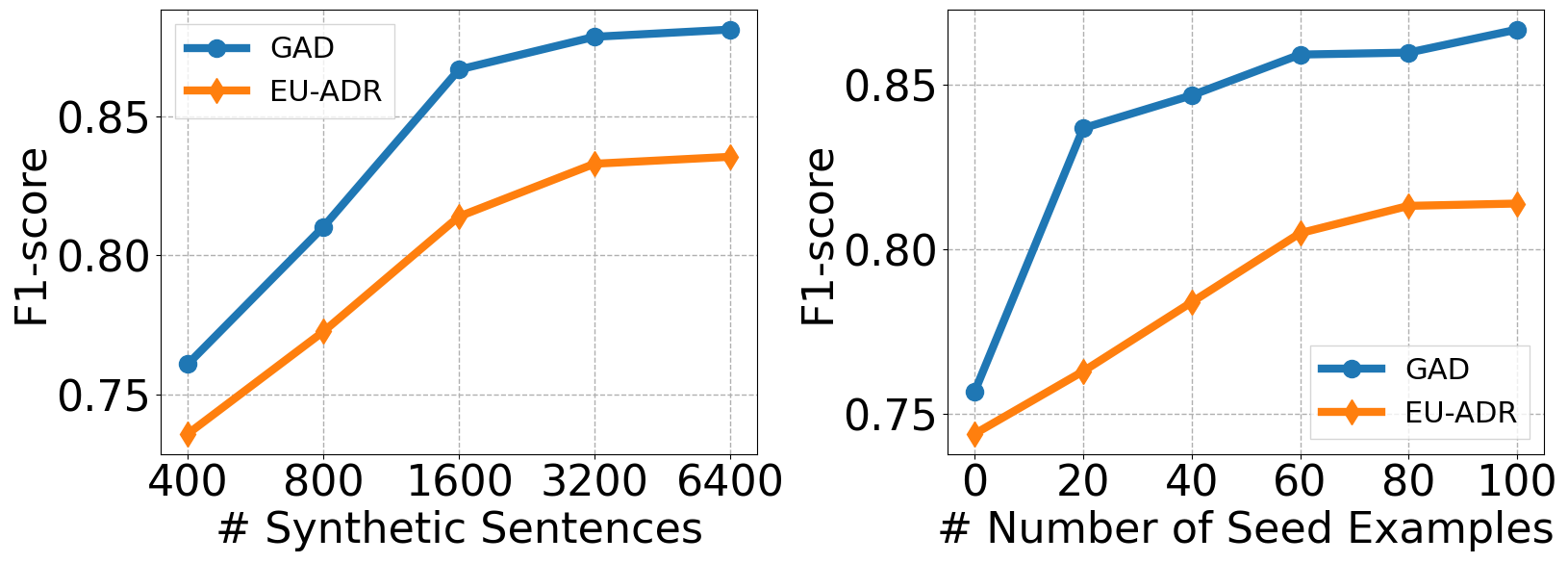}
    \caption{The performance with different numbers of synthetic sentences and the ratio of the seed examples. }\label{fig:effect_num re}
% \end{figure}
\end{wrapfigure}

To investigate the impact of the number of synthetic sentences on the effectiveness of our proposed method, we conducted experiments with varying numbers of synthetic sentences and ratios of seed examples. As mentioned previously, we collected 6437 and 6424 synthetic examples for the GAD and EU-ADR, respectively. In the first experiment, we used a range of $400-6400$ synthetic data to train our local model, while in the second experiment, we varied the pool size of our seed examples from 0 to 100. The results of these experiments are presented in Figure~\ref{fig:effect_num re}. Our findings indicate that increasing the number of synthetic sentences can improve model performance up to a certain point, beyond which the improvement becomes marginal. Specifically, we found that 3500 synthetic sentences are sufficient for obtaining optimal results. Additionally, using a larger number of seed examples can increase the quality and diversity of the generated data. Our experiments demonstrated that 80 seed examples are sufficient for both relation extraction tasks. It is worth noting that not providing seed examples can lead to ChatGPT generating duplicated examples, resulting in a significant drop in model performance.

\section{Analysis of Generated Texts}
\begin{figure}[h]
  \centering
    \includegraphics[width=1.0\textwidth]{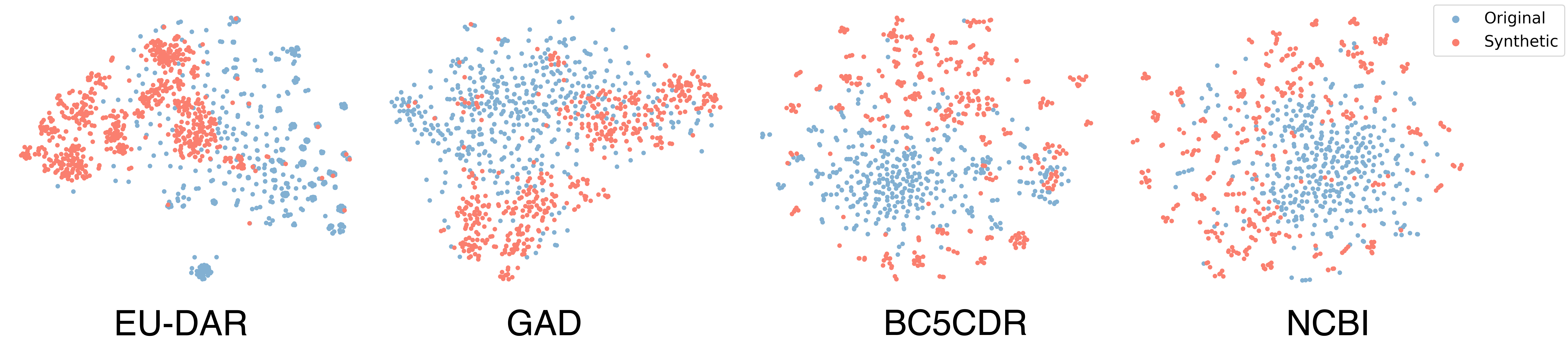}
 \caption{Compare the sentence embedding of the original data and synthetic data.}
 \label{fig: embedding}
\end{figure}

Our previous research has demonstrated that ChatGPT is capable of producing high-quality synthetic data. However, a potential concern is that ChatGPT has been trained on a publicly available dataset, which means that it may have already encountered the dataset used in our experiments. This raises the possibility of ChatGPT inadvertently leaking information from the original dataset. To address this issue, we utilized the sentence transformer to obtain embeddings for both the original and synthetic data, and then projected them using T-SNE. The resulting distribution of sentences, as illustrated in Figure \ref{fig: embedding}, revealed distinct patterns between the synthetic and original data, indicating that ChatGPT did not simply memorize and reproduce the dataset. This distribution shift can also explain the observed performance gap between models fine-tuned on synthetic versus original data. Our future work aims to explore methods for producing synthetic data with a similar distribution as the original data.

\section{Related Work}
In this section, we review the literature related to the topic of our paper, including previous research on Large Language Models and NLP for Biomedical applications..

\textbf{Large Language Models.} Recently, Large Language Models (LLMs) have attracted increasing attention due to their high performance and capability to understand natural language. Consequently, researchers and practitioners are exploring the use of LLMs to assist experts in a variety of domains, including education~\cite{kasneci2023chatgpt}, healthcare~\cite{arora2023promise,singhal2022large}, and content creation~\cite{yuan2022wordcraft}. OpenAI GPT-3~\cite{brown2020language} series has been a breakthrough for LLMs, which is trained by generating the next word in a sequence given the preceding words. These models have been widely used for tasks such as text generation and language modeling, and have also achieved impressive results on many benchmarks. The following works, such as BLOOM~\cite{scao2022bloom}, PaLM~\cite{DBLP:journals/corr/abs-2204-02311} and LLaMA~\cite{touvron2023llama}, are optimized for specific tasks such as code generation and document ranking. Recently, the instruction-tuned version of GPT-3, ChatGPT, has emerged as a game-changer in LLMs. ChatGPT is capable of generating coherent text from scratch. In this work, we use ChatGPT as the zero-shot baseline and we use it to generate syntectic data for us.

\textbf{NLP for Biomedical.} The Natural Language Processing (NLP) technique is widely applied in the biomedical domain, as evidenced by numerous studies~\cite{yandell2002genomics, nedellec2013overview}. NLP for Biomedical has various applications, including the analysis of electronic health records (EHRs)\cite{koleck2019natural, patra2021extracting, ohno2011realizing, wang2009active}, drug discovery\cite{ozturk2020exploring, vamathevan2019applications}, and medical chatbots~\cite{safi2020technical, dharwadkar2018medical}. The use of LLMs for Biomedical is gaining traction among both industry and academic researchers. Previous work~\cite{lee2020biobert, yu2019biobert, zhu2020extracting} has explored the application of NER and RE to biomedical tasks. Biomedical NER and RE has diverse usage in the healthcare domain, including analyzing EHRs~\cite{gorinski2019named, dai2019named, kormilitzin2021med7}, extracting clinical trials~\cite{kang2017eliie, nye2021understanding}, and drug development~\cite{shinozaki2020electronic, rocktaschel2012chemspot, eltyeb2014chemical}. The major challenges facing NLP in Biomedical include developing accurate models for biomedical text analysis and ensuring patient data privacy. In this work, we propose the use of synthetic data to fine-tune offline models, which can not only improve prediction accuracy but also protect patient privacy.

\section{Conclusion}
In this study, we set out to explore the potential of ChatGPT to assist with clinical text mining tasks, with a particular focus on named entity recognition and relation extraction. However, our initial attempts to use ChatGPT directly for these tasks yielded unsatisfactory results and raised privacy concerns. Therefore, we developed a new framework that involved generating high-quality synthetic data with ChatGPT and fine-tuning a local offline model for downstream tasks. The use of synthetic data resulted in significant improvements in the performance of these downstream tasks, while also reducing the time and effort required for data collection and labeling, and addressing data privacy concerns as well. In the future, we aim to further refine our framework to enhance the data quality and extend its application to other clinical tasks.

% References as numbers

\bibliographystyle{vancouver}
\bibliography{amia} 
\end{document}